\pdfoutput=1

\documentclass[11pt]{article}

\usepackage[]{ACL2023}

\usepackage{times}
\usepackage{latexsym}
\usepackage{graphicx}
\usepackage{xcolor}
\usepackage{adjustbox}
\usepackage{multirow}
\usepackage{algorithm}
\usepackage{algpseudocode}
\usepackage{booktabs}
\usepackage{hyperref}
\usepackage{amsmath}
\usepackage{amssymb}
\usepackage{subfig}
\usepackage{caption}

\usepackage{arydshln}
\usepackage{enumitem}
\setlist[itemize]{align=parleft,left=0pt..1em}

\newcommand{\scriptX}{\mathcal{X}}
\newcommand{\scriptY}{\mathcal{Y}}

\newcommand{\scriptA}{\mathcal{A}}
\newcommand{\scriptL}{\mathcal{L}}

\newcommand{\scriptU}{\mathcal{U}}
\newcommand{\scriptT}{\mathcal{T}}

\usepackage[T1]{fontenc}

\usepackage[utf8]{inputenc}

\usepackage{microtype}

\usepackage{inconsolata}

\title{Pseudo Outlier Exposure for Out-of-Distribution Detection using Pretrained Transformers}

\author{Jaeyoung Kim\Thanks{ These authors contributed equally.} \\
  Gachon University \\
  \texttt{kimjeyoung@gachon.ac.kr} \\
  \And
  Kyuheon Jung\footnotemark[1] \\
  Pukyong National University \\ 
  \texttt{kkyuhun94@pukyong.ac.kr} \\
  \And
  Dongbin Na \\
  VUNO, Inc. \\
  \texttt{dongbin.na@vuno.co} \\
  \AND 
  Sion Jang \\
  Alchera Inc.\\
  \texttt{so.jang@alcherainc.com} \\
  \And
  Eunbin Park \\
  Pukyong National University \\
  \texttt{cosmos42@pukyong.ac.kr} \\
  \And
  Sungchul Choi\Thanks{ Corresponding author.} \\
  Pukyong National University \\
  \texttt{sc82.choi@pknu.ac.kr} 
  }

\begin{document}
\maketitle
\begin{abstract}
For real-world language applications, detecting an out-of-distribution (OOD) sample is helpful to alert users or reject such unreliable samples.
However, modern over-parameterized language models often produce overconfident predictions for both in-distribution (ID) and OOD samples.
In particular, language models suffer from OOD samples with a similar semantic representation to ID samples since these OOD samples lie near the ID manifold.
A rejection network can be trained with ID and diverse outlier samples to detect test OOD samples, but explicitly collecting auxiliary OOD datasets brings an additional burden for data collection.
In this paper, we propose a simple but effective method called Pseudo Outlier Exposure (POE) that constructs a surrogate OOD dataset by sequentially masking tokens related to ID classes.
The surrogate OOD sample introduced by POE shows a similar representation to ID data, which is most effective in training a rejection network.
Our method does not require any external OOD data and can be easily implemented within off-the-shelf Transformers.
A comprehensive comparison with state-of-the-art algorithms demonstrates POE's competitiveness on several text classification benchmarks.

\end{abstract}

\section{Introduction}

Pre-trained language models (PLMs) have achieved remarkable success in various natural language processing (NLP) tasks such as question-answering~\cite{yuan-etal-2019-interactive,GPT3}, sentiment analysis~\cite{clark2020electric}, and text categorization~\cite{devlin-etal-2019-bert,XLNET}.
While PLMs have become a de-facto standard promoting classification accuracy, recent studies have found that over-parameterized PLMs often produce overconfident predictions for out-of-distribution (OOD) samples~\cite{MISCALIB1,MISCALIB2}.
For real-world language applications, these unreliable predictions can confuse users when interpreting the model's decisions.
Therefore, language models require the ability to detect OOD samples to instill the reliability in NLP applications. 

The task of detecting OOD samples can be formulated as a binary hypothesis test of detecting whether an input data is from in-distribution (ID) or OOD.
To detect an outlier data, in machine learning communities, the OOD detection task has been studied for many years~\cite{MSP,DE_origin,NLP_dropout_word}. The prior works have proposed effective methods, including post-hoc algorithms~\cite{MAHALANOBIS,DICE}, and training a rejection network by exposing the model to external OOD datasets~\cite{OE}.

However, existing post-hoc methods usually require a subset of actual OOD samples to tune their hyperparameters~\cite{ODIN,REACT},
especially, \citet{generalized_ODIN} find that hyperparameters tuned with limited OOD dataset are not generalized to others.
Thus, these methods are not feasible in real-world applications; moreover, we often cannot know the entire distribution of OOD datasets.
Similarly, training a rejection network not only brings an additional burden for OOD data collection but also may result in sub-par OOD detection performance in deciding which subset of external data to use.
Intuitively, OOD examples that are excessively distant from training samples may not help with OOD detection because easy-to-learn outlier features can be discriminated rather trivially.
Therefore, a desirable trait for OOD samples to effectively train rejection networks is that the OOD sample does not belong to ID but is sufficiently close to the distribution of ID samples~\cite{GAN_outlier}.

In this paper, we primarily focus on detecting OOD samples by constructing a surrogate OOD dataset from training samples rather than using external OOD data to train a rejection network.
To this end, we propose Pseudo Outlier Exposure (POE) which is a procedure to construct a near-OOD set by erasing tokens with high attention scores in training sentences.
A rejection network can then be trained on the training (ID) and constructed OOD datasets to detect OOD samples.
Numerical experiments confirm that our procedure indeed generates surrogate OOD data close to ID examples.
Accordingly, a rejection network trained on this construction outperforms state-of-the-art OOD detection algorithms on several benchmarks.
Our main contributions are:
\begin{itemize}
    \item Our novel method easily constructs a surrogate OOD dataset in an offline manner and can be applied to any ID training data without access to any real OOD sample.
    \item We demonstrate that the resultant surrogate OOD dataset introduced by POE is sufficiently close to the distribution of ID samples, which results in improvement of OOD detection performance for the rejection network.
    \item Through comprehensive comparison with state-of-the-art algorithms, we demonstrate POE's competitiveness on several text classification benchmarks.
\end{itemize}

\section{Related Work}

\subsection{Post-hoc Methods}
Post-hoc methods can detect an OOD sample by manipulating the features or logits of a pre-trained network without changing the weights of the given network.
They have advantages where they do not require re-training a pre-trained classifier to detect OOD samples and can be simply applied in the inference time.
Therefore, post-hoc methods can preserve the classification accuracy for the classifier.
To detect OOD data, \citet{MSP} propose a simple post-hoc algorithm by thresholding the classifier's maximum softmax probability (MSP).
ODIN~\cite{ODIN} adds two additional strategies, temperature scaling and input pre-processing (adding perturbation to the test input) to the MSP for better separating confidence scores between ID and OOD samples.
Treating the distribution of feature vectors of pre-trained models as class-conditional Gaussian distributions, \citet{MAHALANOBIS} suggest the Mahalanobis distance-based confidence scoring rule with statistics of data samples in feature space.
Energy~\cite{ENERGY} propose the OOD scoring rule using an energy score that is aligned with the probability density of the logits of a pre-trained network. They demonstrate that the energy-based scoring rule is less susceptible to the overconfidence issue for a softmax classifier.
ReAct~\cite{REACT} suggests truncating the high activations of the penultimate layer to distinguish distinctive patterns arising when OOD data is fed into the model.
DICE~\cite{DICE} is a sparsification technique that ranks weights by contribution, and then uses the most significant weights to reduce noisy signals in OOD data.

Except for MSP and Energy described above, other methods specify parameter(s) that must be tuned on a reserved OOD subset. However, in many real-world deployment settings, the distribution of entire OOD samples is usually unknown.

\subsection{Training a Rejection Network}

Outlier Exposure (OE; \citealp{OE}) uses auxiliary datasets completely disjoint from the test time data to teach the model a representation for ID/OOD distinctions.
However, in real-world applications, OE has a limitation in that collecting all possible OOD samples is not feasible, and OOD samples may not be known a priori.
$K$-Folden~\cite{kfolden} is an ensemble method that trains $K$ individual classification models. 
Each model is trained on a subset with $K-1$ classes with the remaining class masked unknown (OOD) to the model.
They train each model with a cross-entropy loss for the visible $K-1$ labels and an additional Kullback-Leibler (KL) divergence loss enforcing uniform predictions on the left-one-out label.
For a test time, they simply average the probability distributions produced by these $K$ models and treat the result as the final probability estimate for a test sample.
However, the $K$-Folden lacks scalability to tasks with large classes and requires excessive computational costs because it requires $K$ network instances.
Moreover, their approach cannot be applied to a binary classification task (i.e., $K=2$).

Compared to these studies, our method does not require the actual real-world OOD dataset and only trains a single additional rejection network.

\subsection{Feature Representation Learning}

Contrastive representation learning has shown remarkable performance for both ID classification and OOD detection~\cite{SCL,zhou-etal-2022-knn}.
Compared to the contrastive loss used in self-supervised representation learning~\cite{SIMCLR}, where a model learns the general features of a dataset without labels, \citet{SCL} suggest a supervised-contrastive loss (SCL), instances of the same class form a dense cluster on the model's feature space, whereas instances of different classes are encouraged to be distant from each other.
Motivated by \citet{SCL}, \citet{MCL} propose the margin-based contrastive loss (MCL) to better increase the discrepancy of the representations for ID instances from different classes.
MCL enforces the L2 distances of samples from the same class to be as small as possible, whereas the L2 distances of samples from different classes to be larger than a margin.
They show that the model learned the intra-class compactness achieves advanced OOD detection performance.
Compared to MCL~\cite{MCL} used only $K$ ID classes, we modify MCL by assigning a pseudo OOD set to the ${(K+1)}^{\text{th}}$ OOD class in the contrastive loss.
Thus, our variant version of MCL not only shrinks the manifold of the OOD samples in the feature space but also further maximizes the discrepancy of the representations for ID instances from the surrogate OOD classes.

\section{Method}
Given Transformer-based PLMs with the softmax classifier, we propose a simple but effective method for detecting OOD samples.
We first introduce the proposed method for generating surrogate OOD data and then present a rejection network that is trained with ID and the surrogate OOD.

\noindent \textbf{Notation.} Let $x \in \scriptX_{\text{ID}}$ be a training set, and $y \in \scriptY = \{1,...,K\}$ be a label. For multi-class classification tasks, BERT-style Transformer $f$ can be decomposed by the attention blocks and the last dense layer.
We denote each layer as $f_{\text{att}}$, and $f_{\text{out}}$, respectively. Unless otherwise mentioned, the output of $f_{\text{att}}(\cdot)$ denotes the \verb|[CLS]| feature vector on the last attention block.

\begin{figure}[h]
\centering
\includegraphics[width=7.6cm]{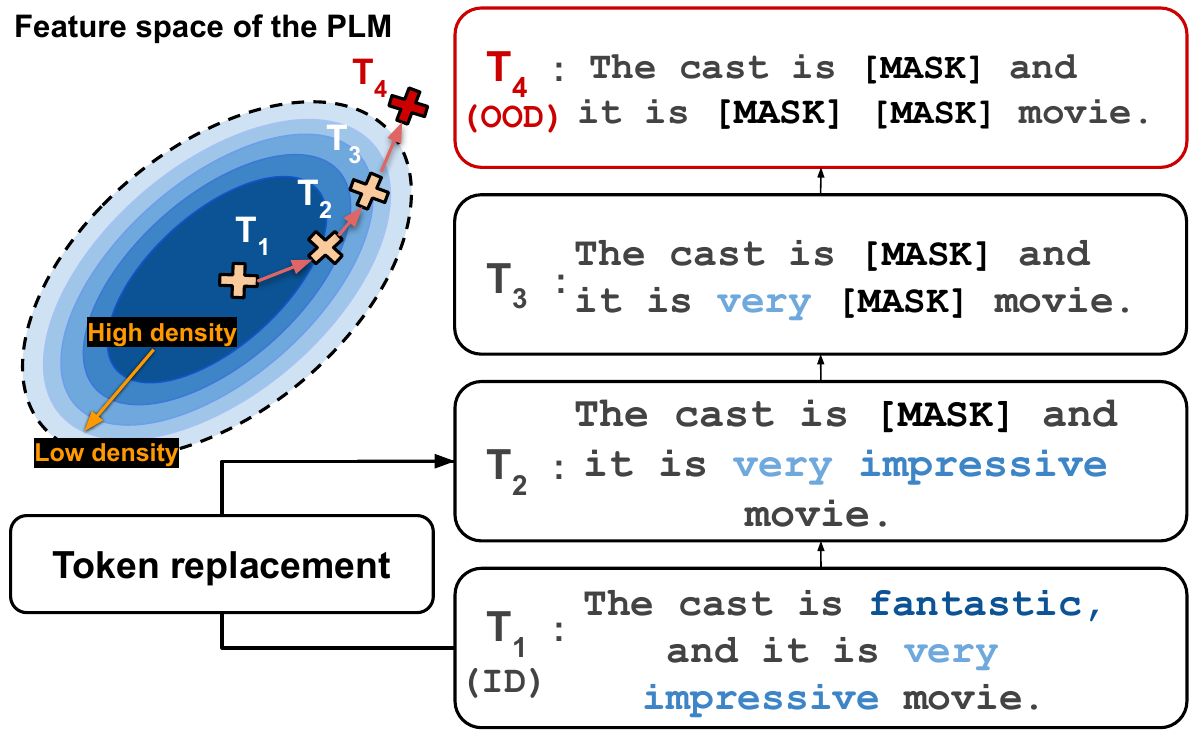}
\caption{An illustration of our surrogate data generation method. In the text boxes, the blue words denote tokens with high attention scores, and the darker words represent higher attention scores than others.}
\label{fig:ood_construction}
\end{figure}

\subsection{Out-of-Distribution Set Construction}

\noindent \textbf{High-level idea.} 
Following \citet{MAHALANOBIS}, we assume that class-conditional features on the PLM's penultimate layer (i.e., the last attention layer) follow the multivariate Gaussian distribution for the training set.
We first calculate the empirical class mean and covariance of the training set.
The former is defined as:
\begin{equation}
    \hat{\mu}_k = \frac{1}{N_k} \sum_{i:y_i=k} f_{\text{att}}(x_i),
\end{equation}
where $N_k$ is the number of samples with class $k$.
The latter can be calculated by:
\begin{equation}
    \hat{\Sigma} = \frac{1}{N} \sum_k \sum_{i:y_i=k} (f_{\text{att}}(x_i) - \hat{\mu}_k)^{\top}(f_{\text{att}}(x_i) - \hat{\mu}_k).
    % \hat{\Sigma} = \frac{1}{N} \sum_c \sum_{i:y_i=c} (f_{\text{att}}(x_i) - \hat{\mu}_c)(f_{\text{att}}(x_i) - \hat{\mu}_c)^{\top}.
\end{equation}

Because our aim is to create surrogate OOD samples that are sufficiently close to the manifold of ID samples, a surrogate OOD sample $\tilde{x}$ would be satisfied the following condition:
\begin{equation}\label{eq:condition}
    \max_{x_i \in \scriptX_{\text{ID}}} M(x_i) < M(\tilde{x}) \leq \min_{x'_i\in \scriptX_{\text{OOD}}} M(x'_i),
    % \max_{x_i \in \scriptX_{\text{ID}}} M(x_{\text{ID}}^i) < M(\tilde{x}) \leq \min_{x_i\in\scriptX_{\text{OOD}}} M(x_{\text{OOD}}^i),  
\end{equation}
where $x' \in \scriptX_{\text{OOD}}$ is an explicit OOD sample (e.g., an OOD sample comes from completely different ID tasks), and $M(\cdot)$ is the Mahalanobis distance between $x$ and the closest class-conditional Gaussian distribution, i.e.,
\begin{equation}\label{eq:mahalanobis}
M(x) = \max_k -(f_{\text{att}}(x) - \hat{\mu}_k)^{\top}\hat{\Sigma}^{-1}(f_{\text{att}}(x) - \hat{\mu}_k).
% M(x) = \max_c -(f_{\text{att}}(x) - \hat{\mu}_c)\hat{\Sigma}^{-1}(f_{\text{att}}(x) - \hat{\mu}_c)^{\top}.
\end{equation}

Considering Eq.~\ref{eq:condition}, we construct the surrogate OOD sample from the training sample, i.e., $x_{\text{train}} \rightarrow \tilde{x}$.  
To obtain the OOD data with a similar semantic representation to ID samples, we gradually erase tokens with high attention scores until $\tilde{x}$ has a larger Mahalanobis distance than the maximum ID Mahalanobis distance.
It can be interpreted that the surrogate OOD sample starts with an ID and gradually turns into OOD as distinct tokens are erased.

\noindent \textbf{Data construction pipeline.}
Let $x = \{x^1, ..., x^S\}$ is the training sample where $S$ is its sequence length, and $x^s$ is the $s^{\text{th}}$ token. 
In the PLM's architecture, we can identify key tokens that mainly affect to the model predictions by leveraging the attention score corresponding to the position of \verb|[CLS]| token.
Using the attention score for each token, we can easily remove these tokens for any training set; thus we construct $\tilde{x}$ excluding tokens that are correlated with ID classes. 

We gradually replace the attention score-based tokens with the \verb|[MASK]| token for $T(\leq S)$ steps using the attention score:
\begin{equation}
\tilde{x}_{t+1}^{s^\ast} \leftarrow \scriptA(\tilde{x}_{t}^{s^\ast}), \,\, t \in \{1, ..., T\},
\end{equation}
where $\scriptA(\cdot)$ is the token replacement function, and $s^\ast$ is the index where the token with the $t^{\text{th}}$ highest attention score is located. 

For each step, we calculate $M(\tilde{x}_t)$, and select $\tilde{x}_{t^{\ast}}$ at the $t^{\ast}$ when $M(\tilde{x}_{t^{\ast}})$ becomes greater than $\max_{i:y_i=k} M(x_{\text{ID}}^i)$. 
For all training samples, we collect the surrogate OOD samples generated by the above process (see Fig.~\ref{fig:ood_construction}).

\subsection{Rejection Network}
The task of detecting OOD samples is a binary hypothesis test
\begin{align}
    f'\left(x\right) = 
    \begin{cases}
        1 & \text{if } x \in \scriptX_{\text{ID}} \\
        0  & \text{if } x \in \scriptX_{\text{OOD}},
    \end{cases}
\end{align}
where $f'$ is a decision model. 
In order for $f'$ to learn the distinctive patterns between ID and OOD samples, we re-train the PLM $f'$ on both ID and the constructed OOD samples by leveraging a supervised contrastive representation learning.
Because we construct the surrogate OOD set, we can explicitly make the model learns distinctive representations of an OOD class as well as $K$ ID classes using the margin-based contrastive loss (MCL; \citealp{MCL}).
Different from MCL, which uses only ID classes, our variant version of MCL contrasts OOD instances to those from different ID classes. 

Let $\{x'_i, y'_i\}_{i=1}^B = \{(x_i, y_i)\, |\, y_i \in \scriptY_{\text{ID}}\}_{i=1}^{B_{\text{I}}} \cup \{\tilde{x}_i, \tilde{y}_i\}_{i=1}^{B_{\text{O}}}$ is a batch of training instances, and $\tilde{y}_i$ is assigned to OOD class $K+1$.
The $B_I$ denotes the size of a batch containing only ID samples, and the $B_O$ denotes the size of a batch containing only our synthesized OOD samples. We denote $A(i)=\{1,...,B\} \backslash \{i\}$ is the set of all anchor instances for the batch samples.

The MCL with $K+1$ classes can be formulated as,
\begin{equation}
\scriptL_{\text{margin}} = \frac{1}{d(B_{\text{I}}+B_{\text{O}})} (\scriptL_{\text{p}} + \scriptL_{\text{n}}),
\end{equation}
where $d$ is the feature dimension of $f_{\text{att}}(x)$, $\scriptL_p$ is the positive loss term that enforces the L2-distances of instances from the same class to be small, and $\scriptL_n$ is the negative loss term that encourages the L2-distances of instances from different classes to be larger than a margin $\xi$. $\scriptL_p$ is calculated by, 
\begin{equation}
\scriptL_{\text{p}} = \sum_{i=1}^B \frac{1}{|P(i)|} \sum_{p\in P(i)} ||f_{\text{att}}(x'_i) - f_{\text{att}}(x'_p) ||^2,
\end{equation}

where $P(i)=\{p\in A(i) | y'_i=y'_p\}$ is the set of indices for the instances from the same class as $y'_i$. The negative loss term is defined as
\begin{equation}\label{eq:neg_mcl}
\small
\scriptL_{\text{n}} = \sum_{i=1}^B \frac{1}{|N(i)|} \sum_{n\in N(i)}\varphi(\xi - ||f_{\text{att}}(x'_i) - f_{\text{att}}(x'_n) ||^2).
\end{equation}

In Eq.~\ref{eq:neg_mcl}, $N(i)=\{n\in A(i) | y'_i \neq y'_n\}$ is the set of indices for the instances from different classes with $y'_i$. $\varphi(\cdot)$ is the ReLU function.
The margin $\xi$ is defined as the maximum distance between positive pairs,
\begin{equation}
\xi = \max_{i=1}^B \max_{p\in P(i)} ||f_{\text{att}}(x'_i) - f_{\text{att}}(x'_p) ||^2.
\end{equation}

In conclusion, we re-train $f'$ with the following objective, $\scriptL_{\text{total}}=\scriptL_{\text{ce}} + \scriptL_{\text{margin}}$, where $\scriptL_{\text{ce}}$ is the cross-entropy loss.
We use $L_{\text{ce}}$ the same as the loss for ID class classification in order to (1) without changing the output node of $f'_{\text{out}}$ and (2) to apply the existing post-hoc methods without modification.

In addition, during re-training, the \verb|[MASK]| token of $\tilde{x}$ is randomly replaced with a word in the PLM's vocabulary so that the model learns about various OOD representations.

\subsection{Out-of-Distribution Scoring Rules}
We use the existing OOD scoring algorithm, which maps the outputs of the model for test samples to OOD detection scores. The low score indicates a low likelihood of being OOD.
Our rejection network can be applied to existing post-hoc methods, and we combine three parameter-free methods with our method in this work.

\begin{itemize}
    \item \textbf{MSP}. \citet{MSP} use the maximum class probability $\max_k \sigma(f(x))$, where $\sigma(\cdot)$ is the softmax function.
    \item \textbf{Energy}~\cite{ENERGY} based scoring rule is defined as $\log \sum_{k=1}^K \text{exp}(f_k(x)) $.
    \item \textbf{Mahalanobis (Maha)}. \citet{MAHALANOBIS} propose the Mahalanobis distance-based scoring rule, but their method requires several hyperparameters should be tuned via a real OOD subset.
    Instead, following \citet{MCL}, we use the parameter-free Mahalanobis distance as a scoring rule: $\max_k -(f_{\text{att}}(x) - \hat{\mu}_k)^{\top}\hat{\Sigma}^{-1}(f_{\text{att}}(x) - \hat{\mu}_k)$.
    Unless otherwise mentioned, we use this scoring rule in our experiments.
\end{itemize}

\section{Setup}
\begin{table}[h]
\centering
\begin{adjustbox}{width=7.6cm,center}
\begin{tabular}{c|cccc|c} \toprule
 \textbf{ID} & \textbf{\# train} & \textbf{\# dev} & \textbf{\# test} & \textbf{\# classes} & \textbf{OOD} \\ \hline
 $\text{CLINC}_{\text{FULL}}$ & 15.0k & 3.0k & 4.5k & 150 & $\text{CLINC}_{\text{OOD}}$ \\
 $\text{CLINC}_{\text{SMALL}}$ & 7.5k & 3.0k & 4.5k & 150 & $\text{CLINC}_{\text{OOD}}$ \\ \hline
 SST2 & 6.2k & 1.5k & 1.8k & 2 & Yelp \\
 Yelp & 448k & 112k & 38k & 2 & SST2 \\ \hline
 $\text{NEWS}_{\text{TOP5}}$ & 51.7k & 0.2k & 17.2k & 5 & $\text{NEWS}_{\text{REST}}$ \\
 IMDB & 20.0k & 5.0k & 25.0k & 2 & c-IMDB \\
\bottomrule
\end{tabular}
\end{adjustbox}
\caption{Data statistics for the six text classification datasets used for our experiments.}
\label{tab:data_statistics}
\end{table}

\subsection{Dataset}
In order to demonstrate the effectiveness of our method, we conduct experiments on common benchmarks for the OOD detection task:

\begin{itemize}
    \item \textbf{$\text{CLINC}_\text{FULL}$} is a user intent classification dataset designed for OOD detection, which consists of 150 intent classes from 10 domains. This dataset
    includes 22.5k ID utterances and 1.2k OOD utterances ($\text{CLINC}_\text{OOD}$).
    \item \textbf{$\text{CLINC}_\text{SMALL}$} is the variant version of the $\text{CLINC}_\text{FULL}$ dataset, in which there are only 50 training utterances per each ID class. This dataset includes 15k ID utterances and 1.2k OOD utterances.
\end{itemize}

Recently, in the field of NLP, \citet{arora-etal-2021-types} categorize OOD samples by two types of distribution shifts: semantic and background shifts.
Because the shifted benchmarks share a common ID text style (background) or content (semantic), these distribution shifts in such near-OOD detection problems are more subtle in comparison to arbitrary ID and OOD dataset pairs (e.g., training and OOD sets come from completely different tasks), and thus, are harder to detect.
We also conduct experiments with semantic shift and background shift benchmarks to verify that POE is effective even with challenging ID/OOD pairs.

The semantic shift benchmark we used is as follows:

\begin{itemize}
    \item \textbf{$\text{NEWS}_{\text{TOP5}}$} is the rebuilt version of the News Category dataset~\cite{misra2018news} for OOD detection. $\text{NEWS}_\text{TOP5}$ contains instances from the five most common classes of the News Category dataset, and the data from the remaining 36 classes are used as OOD ($\text{NEWS}_\text{REST}$). 
    \item \textbf{IMDB}~\cite{maas-etal-2011-learning} is a binary sentiment classification dataset consisting of movie reviews. \citet{kaushik2019learning} construct a set of augmented IMDB samples (c-IMDB) by editing IMDB examples to yield counterfactual labels. As a result, this changes the distribution of semantic features with high a correlation to ID labels. We use the IMDB as ID and c-IMDB as OOD. 
\end{itemize}

For evaluating POE on the background shift, we use the SST2~\cite{socher2013recursive} and Yelp Polarity~\cite{zhang2015character} binary sentiment analysis datasets.
The SST2 consists of movie reviews, whereas the Yelp polarity dataset contains reviews for different businesses, representing a domain shift from SST2.
These datasets are used as ID/OOD pairs (i.e., SST2/Yelp and Yelp/SST2) in our experiments.
The data statistics are described in Tab.~\ref{tab:data_statistics}. 

\subsection{Evaluation Metrics}
The OOD detection performance is measured with respect to the following standard criteria.
\begin{itemize}
    \item \textbf{AUROC} is the area under the receiver operating characteristic curve obtained by varying the operating point. Higher is better.
    \item \textbf{FPR@95TPR (FPR)} is the probability that an OOD (negative) example is classified as a positive when the true positive rate (TPR) is as high as 95\%. Lower is better.

\end{itemize}

\subsection{Training Details}
Two PLMs are used to compare a wide variety of algorithms: BERT-uncased-base~\cite{devlin-etal-2019-bert} and RoBERTa-base~\cite{ROBERTA}. 
The PLMs are optimized with AdamW~\cite{ADAMW}, the weight decay of 0.01, and the learning rate of 2e-5.
We use a batch size of 16 and fine-tune the PLM for 10 epochs on the downstream task.
When training the rejection network, we use $B_{\text{I}}$ of 16, and $B_{\text{O}}$ of 4.
Other training configurations are equal to the above parameters.
For all methods, we report the averaged performance over five runs using different random seeds.
We implement our framework upon Huggingface’s Transformers \cite{wolf-etal-2020-transformers} and implementation codes are available at \url{https://github.com/kimjeyoung/Pseudo_OutlierExposure}.

\subsection{Compared Methods}

\begin{table}[h]
\begin{adjustbox}{width=7.6cm,center}
\begin{tabular}{ccc} \toprule
 & \textbf{Hyperparameter} & \textbf{Range} \\ \hline
\multirow{2}{*}{ODIN} & temperature & \{5, 50, 100, 500, 1000\} \\
 & perturbation & \{0.001, 0.01, 0.1, 1.0\} \\ \hline
DICE & sparsification (\%) & \{10, 30, 50, 90, 99\} \\ \hline
ReAct & truncating (\%) & \{80, 85, 90, 95, 99\} \\ 
\bottomrule
\end{tabular}
\end{adjustbox}
\caption{Hyperparameters for post-hoc methods.}
\label{tab:hyperparams}
\end{table}

We compare our method with six post-hoc methods: MSP, ODIN, Mahalanobis (Maha), Energy, ReAct, and DICE. 
As the orthogonal research, contrastive learning methods that efficiently learn informative feature representations are well-suited for OOD detection. In our work, the recently proposed KNN-Contrastive Learning (KNN; \citealp{zhou-etal-2022-knn}), Supervised-Contrastive Learning (SCL; \citealp{SCL}), and Margin-based Contrastive Learning (MCL; \citealp{MCL}) are also compared.
The hyperparameters of compared contrastive learning methods are followed the original work as possible for a fair comparison.
For the post-hoc methods, excluding parameter-free methods, we report the best OOD detection performance by varying their hyperparameters and adopting their best settings on the test ID/OOD pairs.
The detailed hyperparameter settings are reported in Tab.~\ref{tab:hyperparams}.

\section{Result}
In this section, we present comprehensive experimental evaluations of POE. We compare POE with baselines for CLINC datasets (Sec.~\ref{sec:clinc_result}), followed by empirical results for semantic and background shift benchmarks (Sec.~\ref{sec:shift_result}) and detailed analysis (Sec.~\ref{sec:analysis}).
Due to the space limitation, we evaluate all methods based on RoBERTa in this section, and the experimental results based on BERT are reported in Appendix.

\subsection{Result for CLINC datasets}\label{sec:clinc_result}

\begin{table}[]
\begin{adjustbox}{width=7.7cm,center}
\begin{tabular}{lcccc} \toprule
 & \multicolumn{2}{c}{\textbf{$\text{CLINC}_{\text{FULL}}$}} & \multicolumn{2}{c}{\textbf{$\text{CLINC}_{\text{SMALL}}$}} \\
 & AUROC$\uparrow$ & FPR$\downarrow$ & AUROC$\uparrow$ & FPR$\downarrow$ \\ \hline
 \multicolumn{1}{l|}{MSP} & 95.71 & \multicolumn{1}{c|}{20.08} & 95.18 & 23.91 \\
 \multicolumn{1}{l|}{Energy} & 96.33 & \multicolumn{1}{c|}{15.99} & 95.79 & 19.16 \\
 \multicolumn{1}{l|}{Maha} & 97.55 & \multicolumn{1}{c|}{12.66} & 96.81 & 17.58 \\
\multicolumn{1}{l|}{ODIN} & 96.36 & \multicolumn{1}{c|}{16.49} & 95.73 & 20.24 \\
\multicolumn{1}{l|}{ReAct} & 95.71 & \multicolumn{1}{c|}{20.08} & 95.20 & 23.74 \\
\multicolumn{1}{l|}{DICE} & 95.22 & \multicolumn{1}{c|}{19.16} & 94.35 & 18.08 \\ 
\multicolumn{1}{l|}{KNN-cosine} & 96.37 &  \multicolumn{1}{c|}{19.83}  & 96.03 & 23.91 \\ 
\multicolumn{1}{l|}{KNN-euclidean} & 96.39 &  \multicolumn{1}{c|}{19.33}  & 95.87 & 23.66 \\ 
\hline

\multicolumn{1}{l|}{SCL+MSP} & 94.35 & \multicolumn{1}{c|}{22.91} & 95.89 & 20.40 \\ 
\multicolumn{1}{l|}{MCL+MSP} & 95.73 & \multicolumn{1}{c|}{17.93} & 95.83 & 19.96 \\ 
\multicolumn{1}{l|}{POE+MSP (Ours)} & \textbf{96.58} & \multicolumn{1}{c|}{\textbf{15.58}} & \textbf{96.36}  & \textbf{16.49} \\ 
\hline

\multicolumn{1}{l|}{SCL+Energy} & 95.16 & \multicolumn{1}{c|}{17.08} & 96.50 & 15.24 \\ 
\multicolumn{1}{l|}{MCL+Energy} & 96.41 & \multicolumn{1}{c|}{13.74} & 96.53 & 14.46 \\
\multicolumn{1}{l|}{POE+Energy (Ours)} & \textbf{96.98} & \multicolumn{1}{c|}{\textbf{12.16}} & \textbf{96.62} & \textbf{13.33} \\ 
\hline

\multicolumn{1}{l|}{SCL+Maha} & 97.42 & \multicolumn{1}{c|}{13.91} & 97.06 & 13.24 \\ 
\multicolumn{1}{l|}{MCL+Maha} & 97.63 & \multicolumn{1}{c|}{\textbf{11.24}} & 97.38 & 13.91 \\ 
\multicolumn{1}{l|}{POE+Maha (Ours)} & \textbf{97.66} & \multicolumn{1}{c|}{12.33} & \textbf{97.48} & \textbf{12.08} \\ 

\bottomrule
\end{tabular}
\end{adjustbox}
\caption{Comparison results for the CLINIC datasets. We adopt RoBERTa as a baseline architecture for the experiments. Results are percentages.}

\label{tab:clinc_result}
\end{table}

\begin{table*}[]
\begin{adjustbox}{width=15.9cm,center}
\begin{tabular}{lcccccccccc} \toprule
 & \multicolumn{4}{c}{\textit{\textbf{Background Shift}}} & \multicolumn{4}{c}{\textit{\textbf{Semantic Shift}}} & \multicolumn{2}{c}{\textit{}} \\ \hline
 & \multicolumn{2}{c}{\textbf{SST2}} & \multicolumn{2}{c}{\textbf{Yelp}} & \multicolumn{2}{c}{\textbf{$\text{NEWS}_{\text{TOP5}}$}} & \multicolumn{2}{c}{\textbf{IMDB}} & \multicolumn{2}{c}{\textbf{Average}} \\
 & AUROC$\uparrow$ & FPR$\downarrow$ & AUROC$\uparrow$ & FPR$\downarrow$ & AUROC$\uparrow$ & FPR$\downarrow$ & AUROC$\uparrow$ & FPR$\downarrow$ & AUROC$\uparrow$ & FPR$\downarrow$ \\ \hline
\multicolumn{1}{l|}{MSP} & 67.06 & 92.75 & \multicolumn{1}{|c}{80.81} & 65.29 & \multicolumn{1}{|c}{74.14} & 79.98 & \multicolumn{1}{|c}{59.52} & 92.97 & \multicolumn{1}{|c}{70.38} & 82.75 \\
\multicolumn{1}{l|}{Energy} & 61.53 & 92.99 & \multicolumn{1}{|c}{75.52} & 65.17 & \multicolumn{1}{|c}{75.91} & 75.54 & \multicolumn{1}{|c}{59.11} & 92.56 & \multicolumn{1}{|c}{68.02} & 81.57 \\
\multicolumn{1}{l|}{ODIN} & 67.05 & 92.81 & \multicolumn{1}{|c}{80.80} & 65.32 & \multicolumn{1}{|c}{75.60} & 75.67 & \multicolumn{1}{|c}{59.57} & 92.85 & \multicolumn{1}{|c}{70.76} & 81.66 \\
\multicolumn{1}{l|}{Maha} & 64.64 & 93.53 & \multicolumn{1}{|c}{91.04} & 51.74 & \multicolumn{1}{|c}{79.77} & 68.73 & \multicolumn{1}{|c}{60.48} & 93.96 & \multicolumn{1}{|c}{73.98} & 76.99 \\
\multicolumn{1}{l|}{ReAct} & 67.07 & 92.75 & \multicolumn{1}{|c}{83.21} & 65.30 & \multicolumn{1}{|c}{74.46} & 78.77 & \multicolumn{1}{|c}{59.83} & 92.91 & \multicolumn{1}{|c}{71.14} & 82.43 \\
\multicolumn{1}{l|}{DICE} & 68.49 & 91.30 & \multicolumn{1}{|c}{77.48} & 64.87 & \multicolumn{1}{|c}{74.95} & 84.18 & \multicolumn{1}{|c}{59.88} & 92.54 & \multicolumn{1}{|c}{67.70} & 83.22  \\ 

\multicolumn{1}{l|}{KNN-cosine} & 74.16 & 90.75 & \multicolumn{1}{|c}{79.46} & 65.18 & \multicolumn{1}{|c}{75.17} & 78.52 & \multicolumn{1}{|c}{59.15} & 92.32 & \multicolumn{1}{|c}{71.99} & 81.69 \\ 

\multicolumn{1}{l|}{KNN-euclidean} & 74.48 & 90.63 & \multicolumn{1}{|c}{79.80} & 65.03 & \multicolumn{1}{|c}{75.11} & 78.42 & \multicolumn{1}{|c}{58.54} & 92.50 & \multicolumn{1}{|c}{71.98} & 81.65 \\ 
\hline

\multicolumn{1}{l|}{SCL+MSP} & \underline{59.36} & \underline{94.69} & \multicolumn{1}{|c}{\underline{79.98}} & \underline{70.33} & \multicolumn{1}{|c}{\underline{70.72}} & 79.29 & \multicolumn{1}{|c}{62.34} & 92.20 & \multicolumn{1}{|c}{68.10} & 84.13 \\ 
\multicolumn{1}{l|}{MCL+MSP} & \underline{62.23} & 89.95 & \multicolumn{1}{|c}{89.30} & 58.93 & \multicolumn{1}{|c}{\underline{72.87}} & 77.56 & \multicolumn{1}{|c}{\underline{59.09}} & 92.44 & \multicolumn{1}{|c}{70.87} & 79.72 \\
\multicolumn{1}{l|}{POE+MSP (Ours)} & 70.05 & 91.63 & \multicolumn{1}{|c}{90.47} & 57.94 & \multicolumn{1}{|c}{74.62} & 77.08 & \multicolumn{1}{|c}{\textbf{62.41}} & 92.33 & \multicolumn{1}{|c}{74.39} & 79.75 \\ 
\hline

\multicolumn{1}{l|}{SCL+Energy} & 56.53 & 94.93 & \multicolumn{1}{|c}{76.72} & 70.42 & \multicolumn{1}{|c}{73.69} & 77.02 & \multicolumn{1}{|c}{62.28} & 92.13 & \multicolumn{1}{|c}{67.31} & 83.63 \\ 
\multicolumn{1}{l|}{MCL+Energy} & 61.66 & 89.76 & \multicolumn{1}{|c}{89.17} & 59.00 & \multicolumn{1}{|c}{73.12} & 76.13 & \multicolumn{1}{|c}{58.63} & 92.44 & \multicolumn{1}{|c}{70.65} & 79.33 \\ 
\multicolumn{1}{l|}{POE+Energy (Ours)} & 70.74 & 88.14 & \multicolumn{1}{|c}{90.16} & 57.53 & \multicolumn{1}{|c}{74.31} & 76.07 & \multicolumn{1}{|c}{62.01} & \textbf{92.09}  & \multicolumn{1}{|c}{74.31} & 78.46 \\ 
\hline

\multicolumn{1}{l|}{SCL+Maha} & 75.42 & 82.48 & \multicolumn{1}{|c}{80.88} & 71.34 & \multicolumn{1}{|c}{80.94} & 67.76 & \multicolumn{1}{|c}{61.29} & 93.67 & \multicolumn{1}{|c}{74.63} & 78.81 \\
\multicolumn{1}{l|}{MCL+Maha} & 90.16 & 60.16 & \multicolumn{1}{|c}{97.10} & 17.13 & \multicolumn{1}{|c}{80.19} & 66.21 & \multicolumn{1}{|c}{60.46} & 93.43 & \multicolumn{1}{|c}{81.98} & 59.23 \\ 
\multicolumn{1}{l|}{POE+Maha (Ours)} & \textbf{92.76} & \textbf{36.84} & \multicolumn{1}{|c}{\textbf{97.59}} & \textbf{15.08} & \multicolumn{1}{|c}{\textbf{81.77}} & \textbf{65.50} & \multicolumn{1}{|c}{61.15} & 93.79 & \multicolumn{1}{|c}{\textbf{83.32}} & \textbf{52.80} \\

\bottomrule
\end{tabular}
\end{adjustbox}
\caption{Comparison with state-of-the-art methods. All implementations use RoBERTa.}
\label{tab:shift_result}
\end{table*}

The results in $\text{CLINC}_\text{FULL}$ and $\text{CLINC}_\text{SMALL}$ are presented in Tab.~\ref{tab:clinc_result}, where the best results for each block are highlighted in bold.
Specifically, KNN~\cite{zhou-etal-2022-knn} uses the LOF algorithm~\cite{LOF} as an OOD scoring rule, in which they use two basic distances to calculate the LOF score.   
We denote KNN using Euclidean distance as KNN-euclidean and using cosine distance as KNN-cosine, respectively.

As shown in Tab.~\ref{tab:clinc_result}, POE outperforms all considered baselines on most ID and OOD distribution pairs on CLINC datasets, even though our method never requires access to real OOD data, unlike ODIN, ReAct, and DICE.
Moreover, POE generally performs much better than other contrastive learning methods, especially on the $\text{CLINC}_\text{SMALL}$ which has a small size of training samples (50 instances per class).
This empirical result shows that even if the rejection network is trained with the surrogate OOD set using small number of training samples and it shows the robust performance.

\begin{table}[h]
\begin{adjustbox}{width=7.3cm,center}
\begin{tabular}{l|ccc} \toprule
\textbf{AUROC $\uparrow$} & \textbf{$\text{CLINC}_\text{SMALL}$} & \textbf{SST2} & \textbf{IMDB} \\ \hline
w/o replacement & 96.32 & 91.67 & 61.02  \\ \hline
w/ replacement & 97.48 & 92.76 & 61.15   \\
\bottomrule
\end{tabular}
\end{adjustbox}
\caption{Effect of the replacement technique, which augments the surrogate OOD sample by replacing masked tokens with randomly selected tokens. The OOD detection performance is based on POE+Maha.}
\label{tab:augmentation}
\end{table}

\begin{table}[h]
\begin{adjustbox}{width=6.3cm,center}
\begin{tabular}{l|ccc} \toprule
\textbf{AUROC $\uparrow$} & \textbf{$\text{CLINC}_\text{SMALL}$} & \textbf{SST2} & \textbf{IMDB} \\ \hline
CE & 95.18 & 67.06 & 59.52 \\
CE+KL & 96.90 & 85.68 & 61.01  \\ 
CE+SCL & 97.07 & 92.54 & 60.99  \\ \hline
CE+MCL  & \textbf{97.48} & \textbf{92.76} & \textbf{61.15}   \\
\bottomrule
\end{tabular}
\end{adjustbox}
\caption{Ablation study assessing training objectives. We use the Mahalanobis as an OOD scoring rule.}
\label{tab:ablation}
\end{table}

\subsection{Result for Distribution Shift Benchmarks}
\label{sec:shift_result}

We also conduct the distribution shift experiment using two types of shifted OOD benchmarks to verify that our method can detect the challenging OOD samples successfully.
Tab.~\ref{tab:shift_result} shows OOD detection results for the background and semantic shifts, and the best results are highlighted in bold.

As shown in Tab.~\ref{tab:shift_result}, interestingly, we observe that not only MSP but also the SCL and MCL struggle with these challenging OOD data.
For example, on at least one ID/OOD pair (\textit{underlined entries}), the naive MSP outperforms SCL+MSP and MCL+MSP except for POE+MSP.
In contrast, POE more accurately detects distributionally shifted instances compared to baselines. Especially, POE performs the best with the Mahalanobis distance for both background and semantic shifts.

\subsection{Ablation Study}
Recall that the \verb|[MASK]| token of $\tilde{x}$ is randomly replaced with a word in the PLM's vocabulary for training the rejection network.
We also assess how the replacement technique affects OOD detection performance (see Tab.~\ref{tab:augmentation}).
We observe that using the replacement technique brings additional performance gain by exposing diverse OOD representations to the rejection network.

To investigate the promising design choices of training objectives, we conduct an ablation study by applying each training objective to the rejection network as shown in Tab.~\ref{tab:ablation}.
The CE+KL can be another choice for training the rejection network, which is an additional KL penalty enforcing uniform predictions on the surrogate samples generated by POE, i.e., $\scriptL_{\text{KL}} = KL(f'(\tilde{x}), \scriptU)$, where $\scriptU$ is the uniform distribution over $K$ classes.
Overall, the rejection network is well-suited with a contrastive loss, and CE+MCL shows the best performances for all datasets. 
Different from the KL loss, which can not impose any constraints on the distribution of the rejection network’s inner representation of the given data, the rejection network with the contrastive loss learns the intra-class compactness for both ID and OOD classes, and it further separates the inter-class distances.
We believe that this discriminative feature space introduced by the contrastive loss leads to better OOD detection performance.

\subsection{Analysis}\label{sec:analysis}

\begin{table}[h]
\begin{adjustbox}{width=7.7cm,center}
\begin{tabular}{l|cccc} \toprule
\textbf{Accuracy (\%)} & \textbf{CE} & \textbf{CE+SCL} & \textbf{CE+MCL} & \textbf{CE+POE} \\ \hline
\multicolumn{1}{l|}{$\text{CLINC}_{\text{FULL}}$} & 95.95  & 95.84 & 96.11 & \textbf{96.80} \\
\multicolumn{1}{l|}{$\text{CLINC}_{\text{SMALL}}$} & 95.48  & \textbf{95.99} & 95.66 & 95.91 \\ \hline
\multicolumn{1}{l|}{SST2} & 94.39  & 93.30 & \textbf{94.45} & 93.79 \\
\multicolumn{1}{l|}{Yelp} & 97.75  & 97.76  & 97.65 & \textbf{97.81} \\ \hline
\multicolumn{1}{l|}{$\text{NEWS}_{\text{TOP5}}$} & 92.48  &  92.51 & \textbf{93.04} & 92.49 \\
\multicolumn{1}{l|}{IMDB} & 94.48  & 94.44 & 94.53 & \textbf{94.92} \\ \bottomrule
\end{tabular}
\end{adjustbox}
\caption{ID classification accuracies for contrastive learning methods. }
\label{tab:id_acc}
\end{table}

\begin{table}[h]
\begin{adjustbox}{width=7.0cm,center}
\begin{tabular}{l|ccc} \toprule
\textbf{Target \textbackslash{} ID} & \textbf{$\text{CLINC}_\text{SMALL}$} & \textbf{SST2} &  \textbf{IMDB} \\ \hline
$\text{CLINC}_\text{SMALL}$ & \underline{-3.02} & -7.00 & -22.3 \\
SST2 & -35.96 & \underline{-1.64} & -8.48  \\
Yelp & -33.28 & -2.22 &  -5.82  \\
IMDB & -43.27 & -1.73 & \underline{-1.44}  \\
c-IMDB & -46.42 & -2.07 & -2.58 \\
News & -37.65 & -7.01 & -20.31   \\ \hline
POE $\tilde{\scriptX}$ & -18.86 & -3.25 &  -4.60 \\   
\bottomrule
\end{tabular}
\end{adjustbox}
\caption{Averaged Mahalanobis distance between ID training samples and target datasets.
We report the distance multiplied by $10^{-3}$, and the higher value indicates that the target dataset is closer to ID samples.
We underline values when the target dataset is an ID test set.
}
\label{tab:compare_maha}
\end{table}

\noindent \textbf{Classification Performance}.
When the post-hoc method is applied to the PLM trained on the downstream task, classification accuracy is maintained because its weights do not change.
However, the accuracy may not be preserved when the weights of PLM are fine-tuned using a contrastive loss.

We evaluate the PLM trained with the contrastive loss on the six ID datasets.
The experimental results are shown in Tab.~\ref{tab:id_acc}.
We observe that contrastive losses do not significantly reduce or increase the classification performance, which is similar to the observations by \citet{MCL}.

\noindent \textbf{Analysis of the Surrogate OOD Set}. 
To examine how closely the surrogate OOD samples lie in the ID manifold, we measure the Mahalanobis distance between ID and the surrogate OOD introduced by POE (Tab.~\ref{tab:compare_maha}).
The RoBERTa is trained with the cross-entropy (CE) loss on the ID dataset and we calculate the Mahalanobis distance (Eq.~\ref{eq:mahalanobis}) at the RoBERTa's penultimate layer.
We observe that the surrogate OOD samples produced by POE indeed have similar representations to ID samples.
\begin{figure}[h]
\centering
\includegraphics[width=7.5cm]{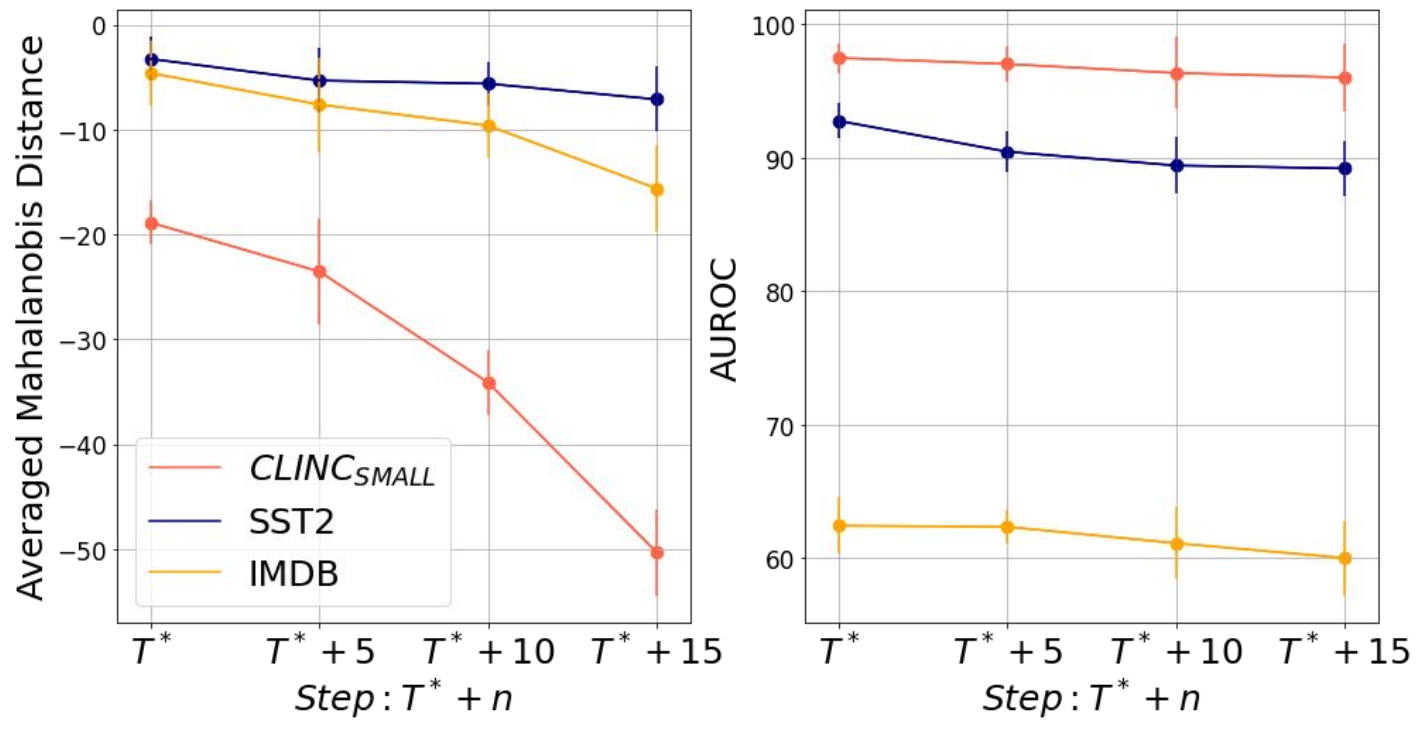}
\caption{POE+Maha’s performances with varying levels of $\scriptT^\ast$. The low Mahalanobis distance implies low similarity between ID and OOD samples.}
\label{fig:varing_t}
\end{figure}
For example, in the feature space of the RoBERTa trained on $\text{CLINC}_{\text{SMALL}}$, the Mahalanobis distance between the surrogate OOD samples and the conditional Gaussian distribution for $\text{CLINC}_{\text{SMALL}}$ has the closest distance to the ID manifold.
For the background (SST2) and semantic shift (IMDB) benchmarks, the IMDB and c-IMDB each has the most similar representation of paired ID set. However, the $\tilde{\scriptX}_{\text{SST2}}$, and $\tilde{\scriptX}_{\text{IMDB}}$ are also sufficiently closed to SST2 and IMDB, respectively.

We also assess whether surrogate OOD samples, which have similar representations to ID samples, are most effective for OOD detection. In our OOD construction, for all training samples, we collect $\tilde{x}_{t^{\ast}} \in \tilde{\scriptX}$ when $M(\tilde{x}_{t^{\ast}})$ becomes greater than $\max_{i:y_i=k} M(x_{\text{ID}}^i)$. Therefore, as $\scriptT^\ast = \{t^\ast_i\}_{i=1}^{N_{\text{Train}}}$ increases, OOD samples that are semantically distant from the ID dataset can be generated.

In Fig.~\ref{fig:varing_t}, we report POE+Maha's OOD detection performances with varying levels of $\scriptT^\ast$.
We identify that surrogate OOD samples produced by a larger $\scriptT^\ast$ further away from the ID samples are generated (Left in Fig.~\ref{fig:varing_t}).
This trait is desirable as ID discriminative tokens are more erased in the surrogate sample.
Moreover, we observe that POE+Maha with surrogate OOD sets introduced by $\scriptT^\ast$ achieves the best AUROC scores for all datasets, whereas the OOD detection performance deteriorates when the reject network is trained with a set of OODs far from the ID.
This empirical result shows that (1) POE leverages the simplicity of erasing attention-based tokens, but it is possible to generate pseudo OOD samples close to the distribution of ID, and (2)
these OOD samples are effective in training the rejection network.

\section{Conclusion}
In this paper, we propose a simple and intuitive OOD construction to train a rejection network. Motivated by the previous observation that OOD samples are most effective when semantically similar to ID samples, POE detects and erases tokens with high attention scores of PLMs. Its resultant surrogate OOD dataset is close to the distribution of ID samples that have been observed to improve the OOD detection performance of the rejection network.
Extensive experiments conducted on challenging ID/OOD pairs show POE's competitiveness.

\section{Limitation}
Although the proposed method achieves significantly improved OOD detection performances compared to the baselines, but POE can not be applied to a naive LSTM, and RNN because our OOD construction is based on an attention score of the PLM.
We leave this issue for future work, but we believe that our proposed method can be used in various NLP tasks as PLMs are now adopted in most fields of NLP tasks. 
While we adopted a masking method using attention scores in this paper, it is not clear that tokens with high attention scores have the most direct impact on the model's predictions~\cite{wiegreffe-pinter-2019-attention}. 
To provide readers with more information, we include additional experimental results in the Appendix to discuss the impact of different masking strategies on OOD detection performance.

\section{Ethics Statement}

The reliability of language models is crucial to the stable deployment of real-world NLP applications. For example, the computer-aided resume recommendation system and neural conversational AI should provide trustworthy predictions because they are intimately related to the issue of trust in new technologies. In this paper, we propose a simple but effective method called POE for OOD detection tasks. We introduce a novel OOD construction pipeline without any external OOD samples to train a rejection network. We hope our work to provide researchers with a new methodological perspective.

\section*{Acknowledgement}
This work was supported by a National Research Foundation of Korea (NRF) grant and funded by the Korean government(No.2021R1C1C1012689 and No.2018R1D1A1B07045825).

\bibliography{anthology,custom}
\bibliographystyle{acl_natbib}

\newpage

\appendix
\section{Additional Result}

\noindent \textbf{Empirical Result for BERT.} We report empirical results for BERT in Table~\ref{tab:bert_appendix_01} and Table~\ref{tab:bert_appendix_02}.

\noindent \textbf{Comparison with other masking strategies.} To provide readers with more information, we compare attention score-based masking with leave-one-out (LOO) method~\cite{wiegreffe-pinter-2019-attention}. In Table~\ref{tab:appendix_03}, both attention-based masking and LOO are effective for the OOD detection task. However, attention-based masking has the advantage of being computationally efficient, as masking priorities can be obtained in a single forward pass. In contrast, LOO is computationally inefficient because it must remove each token in the input sentence one by one to verify the model predictions.

\begin{table*}[h]
\begin{adjustbox}{width=10.0cm, center}
\begin{tabular}{l|cccccc} \toprule
AUROC $\uparrow$ & $\text{CLINC}_{\text{FULL}}$ & $\text{CLINC}_{\text{SMALL}}$ & SST2 & Yelp & News & IMDB \\ \hline
MSP & 96.39 & 95.14 & 65.42 & 83.78 & 71.44 & 56.26 \\
Energy & 97.02 & 96.35 & 62.77 & 82.49 & 73.46 & 54.41 \\
ODIN & 96.98 & 96.01 & 65.43 & 83.70 & 71.52 & 56.25 \\
Maha & 97.18 & 96.81 & 72.87 & 96.79 & 80.35 & 56.40 \\
ReAct & 96.39 & 96.01 & 65.42 & 83.76 & 71.99 & 56.33 \\
DICE & 94.57 & 95.02 & 65.68 & 83.40 & 71.74 & 56.32 \\
KNN-cosine & 96.94 & 96.03 & 72.77 & 84.41 & 73.26 & 57.67 \\
KNN-euclidean & 96.94 & 96.02 & 72.52 & 84.49 & 70.14 & 57.62 \\ \hline 
SCL+MSP & 95.19 & 95.42 & 67.69 & 83.71 & 75.88 & 54.25 \\
MCL+MSP & 95.79 & 95.22 & 63.62 & 84.58 & 75.27 & 58.02 \\
POE+MSP & 96.25 & 95.92 & \textbf{73.27} & 85.24 & 77.23 & 56.64 \\ \hline
SCL+Energy & 96.43 & 96.71 & 65.23 & 81.92 & 76.83 & 56.91 \\
MCL+Energy & 96.69 & 96.82 & 62.71 & 84.96 & 76.03 & 57.67 \\
POE+Energy & \textbf{97.24} & \textbf{96.85} & 72.97 & 85.11 & 77.96 & 56.17 \\ \hline
SCL+Maha & 97.01 & 96.80 & 67.82 & 94.95 & 80.16 & \textbf{59.15} \\
MCL+Maha & 97.18 & 96.81 & 72.88 & 96.79 & 80.35 & 56.40 \\
POE+Maha & 97.03 & 96.19 & 73.03 & \textbf{96.98} & \textbf{80.70} & 57.49 \\ 
\bottomrule
\end{tabular}
\end{adjustbox}
\caption{Comparison results based on BERT. For all methods, we report AUROC (\%) scores. The best results are highlighted in bold.}
\label{tab:bert_appendix_01}
\end{table*}

\begin{table*}[h]
\begin{adjustbox}{width=10.0cm, center}
\begin{tabular}{l|cccccc} \toprule
FPR $\downarrow$ & $\text{CLINC}_{\text{FULL}}$ & $\text{CLINC}_{\text{SMALL}}$ & SST2 & Yelp & News & IMDB \\ \hline
MSP & 17.41 & 20.08 & 91.03 & 69.57 & 82.89 & 91.32 \\
Energy & 12.24 & 14.99 & 97.87 & 68.56 & 76.34 & 91.44 \\
ODIN & 15.39 & 18.22 & 91.01 & 69.47 & 82.83 & 91.35 \\
Maha & 13.91 & 16.08 & 88.77 & 16.97 & 69.31 & 95.48 \\
ReAct & 17.41 & 18.24 & 91.03 & 69.57 & 80.63 & \textbf{91.09} \\
DICE & 22.08 & 20.12 & 91.33 & 68.85 & 82.36 & 91.59 \\
KNN-cosine & 17.33 & 22.75 & 93.64 & 68.21 & 81.57 & 92.85 \\
KNN-euclidean & 17.00 & 22.83 & 93.59 & 68.79 & 86.01 & 92.97 \\ \hline 
SCL+MSP & 22.49 & 20.91 & 90.71 & 68.64 & 77.85 & 92.97 \\
MCL+MSP & 20.91 & 21.58 & 92.64 & 68.05 & 76.77 & 93.02 \\
POE+MSP & 18.89 & 20.24 & 88.69 & 67.91 & 77.38 & 92.76 \\ \hline
SCL+Energy & 14.08 & 15.16 & 95.30 & 68.87 & 72.81 & 92.03 \\
MCL+Energy & 14.91 & 15.66 & 94.08 & 67.92 & 72.15 & 93.02 \\
POE+Energy & \textbf{12.46} & \textbf{14.08} & \textbf{88.52} & 67.36 & 71.16 & 92.14 \\ \hline
SCL+Maha & 12.99 & 16.58 & 89.87 & 27.85 & 71.04 & 94.37 \\
MCL+Maha & 13.91 & 16.08 & 88.77 & 16.97 & \textbf{69.31} & 95.48 \\
POE+Maha & 14.35 & 18.83 & 90.14 & \textbf{16.39} & 70.18 & 94.14 \\ 
\bottomrule
\end{tabular}
\end{adjustbox}
\caption{The OOD detection results based on BERT. Each value indicates the FPR (\%) score.}
\label{tab:bert_appendix_02}
\end{table*}

\begin{table*}[h]
\begin{adjustbox}{width=15.9cm, center}
\begin{tabular}{lccccccccc} \toprule
 & \multicolumn{3}{c}{\textbf{$\text{CLINC}_{\text{SMALL}}$}} & \multicolumn{3}{c}{\textbf{$\text{SST2}$}} & \multicolumn{3}{c}{\textbf{$\text{IMDB}$}}\\
 AUROC$\uparrow$ & \text{MSP} & \text{Energy} & \text{MAHA} & \text{MSP} & \text{Energy} & \text{MAHA} & \text{MSP} & \text{Energy} & \text{MAHA}  \\ \hline
 \multicolumn{1}{l|}{\text{Random masking}} & 94.68 & 95.13 & \multicolumn{1}{c|}{96.32} & 68.43 & 70.28 & \multicolumn{1}{c|}{91.97} & 61.03 & 61.34 & \multicolumn{1}{c|}{55.68} \\
 \multicolumn{1}{l|}{\text{Attention score}} & 96.36 & 96.62 & \multicolumn{1}{c|}{\textbf{97.48}} & 70.05 & \textbf{70.74} & \multicolumn{1}{c|}{\textbf{92.76}} & \textbf{62.41} & \textbf{62.01} & \multicolumn{1}{c|}{61.15} \\
 \multicolumn{1}{l|}{\text{LOO}} & \textbf{96.31} & \textbf{96.84} & \multicolumn{1}{c|}{97.35} & \textbf{71.13} & 70.54 & \multicolumn{1}{c|}{92.50} & 61.88 & 61.71 & \multicolumn{1}{c|}{\textbf{62.26}} \\
\hline

\end{tabular}
\end{adjustbox}

\caption{Comparison result for different masking strategies using RoBERTa. Each value indicates the AUROC (\%) score and the best results are highlighted in bold.}

\label{tab:appendix_03}
\end{table*}

\end{document}